\documentclass{article}

\usepackage{microtype}
\usepackage{graphicx}
\usepackage{subfigure}
\usepackage{booktabs} 

\usepackage{amsmath}
\usepackage{amssymb}
\usepackage{amsthm}

\usepackage{mdframed}
\usepackage{lipsum}

\theoremstyle{definition}
\newmdtheoremenv[style=thmsty]{theo}{Theorem}
\newmdtheoremenv{remark}{Remark}

\newcommand{\innerprod}[2]{\left\langle #1, #2\right\rangle}
\newcommand{\abs}[1]{\left|#1\right|}

\renewcommand{\Pr}{\mathbb{P}}

\renewcommand{\vec}[1]{#1}
\newcommand{\vecSymb}[1]{#1}

\usepackage{hyperref}

\usepackage[accepted]{icml2020}

\icmltitlerunning{Function space locality-sensitive hashing}

\begin{document}

\twocolumn[
\icmltitle{Locality-sensitive hashing in function spaces}

\icmlsetsymbol{equal}{*}

\begin{icmlauthorlist}
\icmlauthor{Will Shand}{equal,ed}
\icmlauthor{Stephen Becker}{equal,ed}
\end{icmlauthorlist}

\icmlcorrespondingauthor{}{}

\icmlaffiliation{ed}{Department of Applied Mathematics, University of Colorado, Boulder}

\icmlkeywords{Machine Learning, Data Science, ICML, LSH, Locality-Sensitive Hashing, Similarity Search, Randomized Algorithms}

\vskip 0.3in
]

\printAffiliationsAndNotice{\icmlEqualContribution} 

\begin{abstract}
We discuss the problem of performing similarity search over function spaces. To perform search over such spaces in a reasonable amount of time, we use {\it locality-sensitive hashing} (LSH). We present two methods 
that allow LSH functions on  $\mathbb{R}^N$ to be extended to $L^p$ spaces:
one using function approximation in an orthonormal basis, and another using (quasi-)Monte Carlo-style techniques. We use the presented hashing schemes to construct an LSH family for Wasserstein distance over one-dimensional, continuous probability distributions.
\end{abstract}

\section{Introduction}

Similarity search over function spaces is an interesting but relatively unexplored problem. The reasons this type of similarity search remains unexplored are fairly straightforward: for one, most datasets encountered in applications are best thought of as consisting of discrete vectors, rather than continuous functions. Even in applications where the data are best modelled as elements of a function space, performing similarity search is very computationally intensive. Calculating just one similarity often requires an integral computation, potentially over a multidimensional domain.

Nonetheless, similarity search over function spaces is more than just a problem of theoretical interest: for instance, Wasserstein metric, which may be defined over a function space, has applications in fields such as image search \cite{10.1109/34.192468}. And an intriguing recent application of similarity search over function spaces is its potential use as a heuristic in optimizing machine learning models; e.g., \cite{AdaBoost-LSH} searches over sets of weak learners generated using AdaBoost. Other applications of function space similarity search in this vein -- e.g., comparing the features learned by neurons in a neural network -- could prove a promising avenue for improving methods to train machine learning models in the future.

Our research seeks to make similarity search over function spaces a tractable problem by accelerating it with locality-sensitive hashing (LSH). In doing so, we can enable a wide range of applications involving similarity search over functions by dramatically reducing computational loads.

\paragraph{Contributions} This paper makes the following contributions:

\begin{itemize}
    \item As a motivation for why we might be interested in performing similarity search in function spaces, we present the example of Wasserstein metric to compare probability distributions. In the case of one-dimensional distributions, we present an algorithm for hashing Wasserstein metrics of order $1 \le p \le 2$.
    \item We discuss how in general hash functions for $\mathbb{R}^N$ can be extended to $L^p$ function spaces. We describe two methods for performing this extension:
    \begin{itemize}
        \item In the specific (but common) case of $p = 2$, we describe a method that uses function approximation in orthonormal bases to perform hashing.
        \item More generally, we use Monte Carlo methods to create an approximate embedding of $L^p_{\mu}(\Omega)$ in $\mathbb{R}^N$ in order to perform hashing. This method works for all $p > 0$.
    \end{itemize}
\end{itemize}

\paragraph{Notation} In this paper, we use $\Omega$ to denote a (measurable) subset of $\mathbb{R}^n$. $L^p_{\mu}(\Omega)$ signifies the $L^p$ function space, defining $\|f\|_{L^p_{\mu}} \equiv \left(\int_{\Omega} \abs{f(x)}^p d\mu(x)\right)^{1/p}$ (which is a norm if $p\ge 1$), over the measure space $(\Omega,\mathcal{A},\mu)$, where $\mathcal{A}$ is a $\sigma$-algebra on $\Omega$. $L^p(\Omega)$ is used to imply that the measure $\mu$ is Lebesgue measure. In the special case of $p = 2$, the inner product is denote $\innerprod{\cdot}{\cdot}_{L^2_{\mu}}$ (and $\Omega$ is implicit).

Additionally, we let $\ell^p(I)$ be the space of real sequences indexed by some set $I$, whose norm is defined as $\|\vec{x}\|_{\ell^p(I)} \equiv \left(\sum_{i\in I} \abs{x_i}^p\right)^{1/p}$; $\ell^p_N$ is shorthand for $\ell^p(\{1,\ldots,N\})$. The inner product for the case of $p = 2$ is denoted $\innerprod{\cdot}{\cdot}_{\ell^2}$ (and $N$ is implicit).

Finally, $W^p(f,g)$ is used to indicate the order-$p$ Wasserstein distance between probability distributions $f$ and $g$. The Wasserstein metric is defined in Section \ref{subsec:wasserstein-metric-definition}.

\section{Background}

\subsection{Locality-sensitive hashing}
{\it Locality-sensitive hashing} is a method for accelerating similarity search (e.g. via $k$-nearest neighbors) that uses hash tables to reduce the number of queries that must be performed. Roughly, a family of hash functions $h:X\to\mathbb{Z}$ is {\it locality-sensitive} for some similarity function $s:X\times X\to\mathbb{R}$ if a hash function, drawn at random from the family, maps sufficiently similar inputs to the same hash with high probability, while keeping a small probability of a hash collision for sufficiently disparate inputs.

The idea behind LSH is that given a query with which to perform similarity search, we can reduce the size of our search space by only comparing our query against those elements of the database that experience a hash collision with the query. This can accelerate the process of performing similarity search by orders of magnitude, especially when our database is large. 

To fine-tune the probability of a hash collision, one generally uses multiple hash functions sampled from the same LSH family simultaneously, replacing a single hash with a tuple of hashes. It is also common practice to use multiple hash tables simultaneously, so that a hash collision between a query and a database entry in one table is equivalent to a hash collision in every table. With {\it multi-probe LSH} \cite{multi-probe-lsh}, one can further fine-tune collision probabilities by looking through buckets that don't necessarily correspond exactly to the query point's hash, but rather correspond to ``nearby'' hashes.

LSH families exist for a number of different similarity measures. Of interest in this paper are LSH families used for comparing vectors in $\mathbb{R}^N$, such as hash functions for cosine similarity \cite{SimHash}, $\ell^p$ distance for all $p\in(0,2]$ \cite{LpHash}, and inner product similarity \cite{MIPSHash,SignALSH}.

\subsection{Wasserstein metric} \label{subsec:wasserstein-metric-definition}
The $p$-Wasserstein distance is a metric between a pair of probability distributions $f$ and $g$ on a metric space $(\Omega,d)$. It is defined as
\begin{equation} \label{eq:wasserstein-distance-generic}
    W^p(f,g) = \inf_{\gamma\in\Gamma(f,g)}\left(\int_{\Omega\times\Omega} d(x,y)^p \hspace{0.15cm}d\gamma(f,g)\right)^{1/p}
\end{equation}

where $\Gamma(f,g)$ is the set of probability distributions on $\Omega\times\Omega$ with marginals $f$ and $g$. 

There is also a discrete analogue to (\ref{eq:wasserstein-distance-generic}) for $\vec{m}_a,\vec{m}_b\in\mathbb{R}^n$ representing discrete probability distributions over a set of $n$ points, where $W^p(\vec{m}_a,\vec{m}_b)$ is formulated as the solution to the linear program:
\begin{align} \label{eq:discrete-wasserstein-distance}
    W^p(\vec{m}_a,\vec{m}_b) & = \min_f \sum_{i=1}^n\sum_{j=1}^n f_{i,j} d^p_{i,j} \\
    \nonumber \text{s.t.} & \hspace{0.2cm} f_{i,j} \ge 0 \hspace{0.5cm} \forall i,j
\end{align}
\begin{align*}
    \nonumber \hspace{0.2cm} \sum_{i=1}^n f_{i,j} = \vec{m}_{bj} \hspace{0.3cm} \forall j,  \hspace{1cm} & \sum_{j=1}^n f_{i,j} = \vec{m}_{ai} \hspace{0.3cm} \forall i \\
    \hspace{0.2cm} \sum_{i=1}^n\sum_{j=1}^n f_{i,j} & = 1
\end{align*}

where $d_{i,j}$ is the distance between points $x_i$ and $x_j$, and  $f_{i,j}$ is viewed as the ``flow of mass" from $x_i$ to $x_j$, analogous to $\gamma$ in \eqref{eq:wasserstein-distance-generic}. Equations (\ref{eq:wasserstein-distance-generic}) and (\ref{eq:discrete-wasserstein-distance}) are both deeply tied to optimal transport problems --- in particular, $W^1$ is often referred to as the {\it earth mover's distance}. 

For this paper, we will consider 
the special but nonetheless useful case of equation (\ref{eq:wasserstein-distance-generic}) where $\Omega \subseteq \mathbb{R}$ and $d(x,y) = \abs{x - y}$.
In this case, 
the Wasserstein distance has the closed-form expression
\begin{align} \label{eq:wasserstein-distance-1d}
    \nonumber W^p(f,g) & = \left(\int_0^1 \abs{F^{-1}(x) - G^{-1}(x)}^{p} \hspace{0.15cm} dx\right)^{1/p} \\
    & = \left\|F^{-1} - G^{-1}\right\|_{L^p}
\end{align}
where $F$ and $G$ are the c.d.f.'s of $f$ and $g$, respectively. This is valid for any $p\ge 1$ due to convexity of  norms~\citep[Prop 2.17]{santambrogio2015optimal}

While computing $W^p(f,g)$ via equation \eqref{eq:wasserstein-distance-generic} is typically  expensive, the simplified expression for one-dimensional Wasserstein distance is significantly more tractable. Nonetheless, it can still present computational problems for similarity search. For one, calculating \eqref{eq:wasserstein-distance-1d} with quadrature rules can be expensive when we want to achieve low error. It is also often the case that we don't have explicit representations for $f$ and $g$, but rather samples of the underlying random variables $X_f$ and $X_g$ with those distributions. Approximating $\|F^{-1} - G^{-1}\|_{L^p}$ in this case is difficult using quadrature rules, especially if the number of samples of each random variable is different. The easiest way to approximate $W^p$ is to model $F^{-1}$ and $G^{-1}$ as step functions, but this approach may have relatively high numerical error. Moreover, computing $W^p$ still takes at least $O(m + n)$ time (where $m$ and $n$ are the number of samples of $X_f$ and $X_g$ respectively), which can be painfully large if we have many samples of at least one of the random variables.

LSH offers a promising method to accelerate similarity search with Wasserstein distance. Our goal in Section \ref{sec:methods} is to identify methods by which we can construct LSH families for similarities defined over function spaces, including 1D Wasserstein distance.

\begin{remark}
If we can construct an LSH family for the similarity function $s(f,g) = \|f - g\|_{L^p}$ on the space $L^p([0,1])$, then it is apparent from equation \eqref{eq:wasserstein-distance-1d} that we can apply a function from that LSH family to $F^{-1}$ and $G^{-1}$ to get a locality-sensitive hash for $W^p$.

In Section \ref{sec:methods} we will provide two methods for extending the $\ell^p$-distance hash of \citet{LpHash} to $L^p_{\mu}(\Omega)$. Since the original $\ell^p$ hash works for all $p\in(0,2]$, and Equation \eqref{eq:wasserstein-distance-1d} applies to all $p \ge 1$, we will have an LSH family for all $W^p$ distances with $1 \le p \le 2$.
\end{remark}

\subsection{Related work}
There have previously been attempts to construct LSH families for the discrete analogue to Wasserstein distance in equation \eqref{eq:discrete-wasserstein-distance}, although we are not aware of any attempts to use LSH for the continuous problem \eqref{eq:wasserstein-distance-generic}. \citet{SimHash} applied a transformation to $f$ and $g$ such that the distance between them was bounded below by $W^1(f,g)$ and above by $O(\log{n}\log{\log{n}})W^1(f,g)$. \citet{FastImageRetrieval} presented a technique for approximately embedding $W^1$ in $\ell^1_N$, over which they then constructed an LSH family. More generally, the $\ell^p$-distance hash of \citet{LpHash} can be used to generate an LSH family for any similarity function or metric space, provided that the similarity or metric can be approximately embedded in $\ell^p_N$ for $p\in(0,2]$.

Research into LSH for function spaces is relatively sparse. \citet{LSH-Probability-Distributions}, covering a hash function for cosine similarity between probability distributions, seems to be the clearest example of LSH over spaces of functions. \citet{AdaBoost-LSH} also handles locality-sensitive hashing in function spaces by using LSH over the weak learners generated using AdaBoost. In both papers, the hash functions that are presented are fairly restrictive and only apply in some unique circumstances. In contrast, this paper contributes two different methods for constructing locality-sensitive hash functions on many different measures of similarity, over a much larger class of function spaces.

\section{Methods} \label{sec:methods}
Our general approach is to create an embedding $T: L_{\mu}^p(\Omega) \to \ell^p_N$ that preserves the distance between functions with minimal distortion. After achieving this, LSH functions for a variety of similarities (e.g. $L^p$ distance and cosine similarity) can be used to hash functions $f\in L^p_{\mu}(\Omega)$ by hashing $T(f)$.

We present two methods in this vein that can be used to extend LSH functions for $\ell^p_N$ to function spaces:

\begin{itemize}
    \item In the special case of $p = 2$, we hash $L^2_{\mu}(\Omega)$ by approximating functions in an orthonormal basis in quasilinear time.
    \item For the more general case of {\it all} $p > 0$ (including the case $p = 2$ where one does not have a sufficiently convenient orthonormal basis for the domain $\Omega$ and measure $\mu$), we use (quasi-)Monte Carlo methods to embed $L^p_{\mu}(\Omega)$ in $\mathbb{R}^N$. We achieve $O\left(\frac{1}{\sqrt{N}}\right)$ or $O\left(\frac{1}{N}\right)$ error in time linear in $N$.
\end{itemize}

In both methods, the embedding $T(\cdot)$ has error inversely correlated with $N$, with the guarantee that as $N\to\infty$ the error converges to zero. We will see that $N$ can be increased as needed, so that we can achieve arbitrarily small error in our embeddings (and hence better hash functions).

\subsection{Approximation in an orthonormal basis} \label{subsec:approxfun-approach}
Start by considering the case of hashing elements of $L^2_{\mu}(\Omega)$. If $\{e_i\}_{i\in\mathbb{N}}$ is an orthonormal basis for $L^2_{\mu}(\Omega)$ (e.g. a wavelet basis), then the mapping from $L^2_{\mu}(\Omega)$ to $\ell^2(\mathbb{N})$ given by $f \mapsto (\innerprod{e_1}{f}_{L_{\mu}^2},\innerprod{e_2}{f}_{L_{\mu}^2},\ldots)$ is a Hilbert space isomorphism between between $L^2_{\mu}(\Omega)$ and $\ell^2(\mathbb{N})$.

Suppose we truncate this mapping for a function $f$ after $N_f$ terms. If $N_f$ is sufficiently large, then we have the approximation (to be made precise later)
\begin{equation*}
    f(x) \approx \hat{f}(x) \equiv \sum_{i=1}^{N_f} \innerprod{e_i}{f}_{L_{\mu}^2}e_i(x)
\end{equation*}

where the right-hand side of the equation above is close to $f(x)$ in $L^2_{\mu}$-norm. Now let $N$ be some integer greater than $N_f$, and define $T:L^2_{\mu}\to\ell^2_N$ as
\begin{equation} \label{eq:L2-l2-transformation}
    T_N(f) = \left(\innerprod{e_1}{f}_{L^2_{\mu}}, \ldots, \innerprod{e_{N_f}}{f}_{L^2_{\mu}}, 0, \ldots, 0\right).
\end{equation}

Then $T_N$ approximately preserves $\|f - g\|_{L^2_{\mu}}$ and $\innerprod{f}{g}_{L^2_{\mu}}$ in the case that $N_g \le N$:
\begin{align*}
    \|f - g\|_{L^2_{\mu}}^2 & \approx \|\hat{f} - \hat{g}\|_{L^2_{\mu}}^2 = \|T_N(f) - T_N(g)\|_{\ell^2_N} \\
    \innerprod{f}{g}_{L^2_{\mu}} & \approx \innerprod{\hat{f}}{\hat{g}}_{L^2_{\mu}} = \innerprod{T_N(f)}{T_N(g)}_{\ell^2_N}
\end{align*}

As long as 
we choose $N \ge N_f$ for all functions $f$ in our dataset,
then $T_N$, as defined above, is an approximate embedding of $L^2_{\mu}$ in $\ell^2_N$.

\paragraph{Using orthonormal bases to compute hashes} To hash $L^2_{\mu}(\Omega)$, we first map $f\in L^2_{\mu}(\Omega)$ to $T(f) \equiv T_N(f)$, and then apply a locality-sensitive hash on $\ell^2_N$ for whatever similarity we are interested in. In theory, $N$ may be extremely large; however, we can use the fact that $T(f)$ is zero in its last $N - N_f$ coefficients to significantly accelerate the process of hashing $T(f)$.

As an example, we will consider the case of extending the $\ell^2$-distance hash of \citet{LpHash} to hashing $L^2_{\mu}(\Omega)$. This hash is computed for a vector in $\ell^2_N$ as
\begin{equation} \label{eq:lphash-definition}
    h(\vec{x}) = \left\lfloor\frac{\vecSymb{\alpha}^{\top}\vec{x}}{r} + b\right\rfloor
\end{equation}

where $\vecSymb{\alpha}\in\mathbb{R}^N$ has i.i.d. entries randomly sampled from the standard normal distribution $\mathcal{N}(0,1)$, $b\sim\text{Uniform}([0,1])$, and $r$ is a positive number chosen by the user. To extend this hash to $L^2_{\mu}$, we will simply hash $h(T(f))$, since $T(f)\in\ell^2_N$.

Instead of generating all $N$ coefficients of $\vecSymb{\alpha}$ --- which may require a massive amount of memory, and requires us to place an upper bound on $N_f$ --- we lazily generate new coefficients of $\vecSymb{\alpha}$ when we encounter a new input $f$ for which $N_f$ is greater than the length of $\vecSymb{\alpha}$. This approach is used in the pseudocode shown in Algorithm \ref{alg:L2Hash-approxfun}, which demonstrates the construction and usage of a locality-sensitive hash function for $L^2_{\mu}$ distance.

\begin{remark}
The sparsity pattern of $T(f)$ makes computing dot products like $\innerprod{\vecSymb{\alpha}}{T(f)}_{\ell^2_N}$ efficient. Since $T(f)$ is zero in its last $N - N_f$ coefficients, $\innerprod{\vecSymb{\alpha}}{T(f)}_{\ell^2_N} = \sum_{i=1}^{N_f} \alpha_i \cdot \left[T(f)\right]_i$.

In addition, this sparsity pattern also means that we never need to know the full vector $\vecSymb{\alpha}$. Instead, we can just append new randomly generated coefficients to $\vecSymb{\alpha}$ when we encounter a new largest value of $N_f$.
\end{remark}

\begin{algorithm}[tb]
   \caption{Function LSH for $L^2_{\mu}$ distance, based on \cite{LpHash}, using function approximation}
   \label{alg:L2Hash-approxfun}
\begin{algorithmic}
    \STATE {\bfseries Input:} function $f$, integer $N_f$, orthonormal basis $\{e_i\}_{i\in\mathbb{N}}$, coefficients $\{\alpha_i\}_{i=1}^N$, and parameters $b$ and $r$
    \STATE {\bfseries Output:} a signed-integer hash $h$ and coefficients $\{\alpha_i\}_{i=1}^{\max{(N,N_f)}}$
    \STATE
    \STATE $\gamma \gets \left(\innerprod{e_1}{f},\ldots,\innerprod{e_{N_f}}{f}\right)$
    \IF{$N_f > n$}
    \STATE Sample $\alpha_{N+1},\ldots,\alpha_{N_f}$ i.i.d. from $\mathcal{N}(0,1)$
    \STATE $N \gets N_f$
    \ENDIF
    \STATE $h \gets \left\lfloor \left(\sum_{i=1}^{N_f} \alpha_i\gamma_i\right)/r + b \right\rfloor$
    \STATE {\bfseries return } $h$, $(\alpha_1,\ldots,\alpha_N)$
\end{algorithmic}
\end{algorithm}

\paragraph{Error analysis} Let $\varepsilon_f(x) = f(x) - \hat{f}(x)$ and $\varepsilon_g(x) = g(x) - \hat{g}(x)$ be the errors made by approximating $f$ and $g$ with a finite number of basis elements.
We have the following bounds on the error induced by the embedding $T_N$ of $L^p_{\mu}(\Omega)$ in $\ell^p_N$:
\begin{align}
    \bigg|\|f - g\|_{L^2_{\mu}} - \|T_N(f) - T_N(g)\|_{\ell^2}\bigg| & = \notag \\ \nonumber \abs{\|f - g\|_{L^2_{\mu}} - \|\hat{f} - \hat{g}\|_{L^2_{\mu}}} & \le \|\varepsilon_f - \varepsilon_g\|_{L^2_{\mu}} \\
    \nonumber & \le \|\varepsilon_f\|_{L^2_{\mu}} + \|\varepsilon_g\|_{L^2_{\mu}}
\end{align}
and 
\begin{align}
    &\abs{\innerprod{f}{g}_{L^2_{\mu}} - \innerprod{T(f)}{T(g)}_{\ell^2} }
    = \abs{\innerprod{f}{g}_{L^2_{\mu}} - \innerprod{\hat{f}}{\hat{g}}_{L^2_{\mu}}} \notag \\
    &\;= \nonumber \abs{\innerprod{\varepsilon_f}{g}_{L^2_{\mu}} + \innerprod{f}{\varepsilon_g}_{L^2_{\mu}} + \innerprod{\varepsilon_f}{\varepsilon_g}_{L^2_{\mu}}} \\
    &\;\le \nonumber \|f\|_{L^2_{\mu}}\cdot\|\varepsilon_g\|_{L^2_{\mu}} + \|g\|_{L^2_{\mu}}\cdot\|\varepsilon_f\|_{L^2_{\mu}} + \|\varepsilon_f\|_{L^2_{\mu}}\cdot\|\varepsilon_g\|_{L^2_{\mu}}
\end{align}

In other words, if $N_f$ and $N_g$ are chosen such that $\|\varepsilon_f\|_{L^2_{\mu}}$ and $\|\varepsilon_g\|_{L^2_{\mu}}$ are both size $O(\varepsilon)$ for some  $\varepsilon > 0$, then the absolute error of $T_N$ in approximating $\|f - g\|_{L^2_{\mu}}$ is $O(\varepsilon)$. Meanwhile, the absolute error in approximating $\innerprod{f}{g}_{L^2_{\mu}}$ is $O\left[\left(\|f\|_{L^2_{\mu}} + \|g\|_{L^2_{\mu}} + \varepsilon\right)\varepsilon\right]$.

We will use the $\ell^p$-distance hash from \citet{LpHash} as an example for how this error can impact the probability of hash collision. The hash collision probability presented in that paper for two inputs $\vec{x}$ and $\vec{y}$ is
\begin{align*}
    \Pr[h(\vec{x}) = h(\vec{y})] & = \int_0^r \frac{1}{c}f_p\left(\frac{t}{c}\right)\left(1 - \frac{t}{r}\right) \hspace{0.15cm} dt \\
    & = \int_0^{r/c} f_p(s) \left(1 - \frac{cs}{r}\right) \hspace{0.15cm} ds
\end{align*}

where $r$ is a user-defined parameter, $c = \|\vec{x} - \vec{y}\|_p$, and $f_p$ is the p.d.f.\ of the absolute value of the underlying $p$-stably distributed random variable.

Suppose that $N_f$ and $N_g$ are such that $\|\varepsilon_f\|_{L^2_{\mu}}$ and $\|\varepsilon_g\|_{L^2_{\mu}}$ are both $\le \varepsilon/2$. Let $H(f) = h(T_N(f))$ (where $h$ is the $\ell^p$-distance hash function for $\ell^p_N$) and let $c = \|f - g\|_{L^2_{\mu}}$. Then we have the following bounds on the probability of a hash collision between $f$ and $g$:

\begin{theo} \label{thm:approxfun-lphash-prob-bounds}
The hash collision probability is bounded above by
\begin{equation*}
    \mathbb{P}\left[H(f) = H(g)\right] \le P + \min{\left(\frac{\varepsilon}{c - \varepsilon}, \frac{\varepsilon r\|f_p\|_{L^{\infty}}}{2(c-\varepsilon)^2}\right)}
\end{equation*}
and below by
\begin{equation*}
    \mathbb{P}[H(f) = H(g)] \ge P - \min{\left(\frac{2\varepsilon}{c+\varepsilon}, \frac{\varepsilon r\|f_p\|_{L^{\infty}}}{2(c+\varepsilon)^2}\right)}
\end{equation*}
where $P = \int_0^{r/c} f_p(s)\left(1 - \frac{cs}{r}\right) \hspace{0.15cm} ds$ is the collision probability when $\varepsilon = 0$.
\end{theo}

\paragraph{Proof of Theorem \ref{thm:approxfun-lphash-prob-bounds}} The proof of these inequalities is just an application of H\"older's theorem. Since $\Pr[h(x) = h(y)]$ is monotone decreasing in $c$, the hash collision probability is bounded above by $\Pr[h(x) = h(y)]$ for $c - \varepsilon$, and below by $\Pr[h(x) = h(y)]$ for $c + \varepsilon$. Thus we have the upper bound

\begin{align*}
    & \int_0^{r/(c-\varepsilon)} f_p(s)\left(1 - \frac{(c-\varepsilon)s}{r}\right) \hspace{0.15cm} ds \\
    & = \int_0^{r/(c-\varepsilon)} \left(f_p(s)\left(1 - \frac{cs}{r}\right) + \frac{s\varepsilon f_p(s)}{r}\right) \hspace{0.15cm}ds \\
    & \le P + \frac{\varepsilon}{r}\int_0^{r/(c-\varepsilon)} sf_p(s) \hspace{0.15cm} ds \\
    & \le P + \frac{\varepsilon}{c - \varepsilon}
\end{align*}

and the lower bound

\begin{align*}
    & \int_0^{r/(c+\varepsilon)} f_p(s)\left(1 - \frac{(c+\varepsilon)s}{r}\right) \hspace{0.15cm} ds \\
    & = \int_0^{r/(c+\varepsilon)}\left[f_p(s)\left(1 - \frac{cs}{r}\right) - \frac{s\varepsilon f_p(s)}{r}\right] \hspace{0.15cm} ds \\
    & = P - \frac{\varepsilon}{r}\int_0^{r/(c+\varepsilon)} sf_p(s) \hspace{0.05cm}ds - \int_{r/(c+\varepsilon)}^{r/c} f_p(s)\left(1 - \frac{cs}{r}\right)\hspace{0.05cm}ds \\
    & \ge P - \frac{\varepsilon}{c+\varepsilon} - \int_{r/(c+\varepsilon)}^{r/c} f_p(s)\left(1 - \frac{cs}{r}\right)\hspace{0.15cm}ds \\
    & \ge P - \frac{\varepsilon}{c + \varepsilon} - \left(1 - \frac{c}{c+\varepsilon}\right) \\
    & = P - \frac{2\varepsilon}{c+\varepsilon}
\end{align*}

To compute the upper bound, we used the inequality 
\begin{equation*}
\int_0^{r/(c-\varepsilon)} sf_p(s) \hspace{0.15cm} ds \le \left(\sup_{s\in[0,r/(c-\varepsilon)]} s\right)\|f_p\|_{L^1} = \frac{r}{c-\varepsilon}
\end{equation*}
which is a result of H\"older's inequality and the fact that $f$ is a probability distribution function. If we instead use H\"older's inequality as
\begin{equation*}
    \int_0^{r/(c-\varepsilon)} sf_p(s) \hspace{0.15cm} ds \le \|f_p\|_{L^{\infty}}\int_0^{r/(c-\varepsilon)} s \hspace{0.15cm} ds = \frac{r^2\|f_p\|_{L^{\infty}}}{2(c-\varepsilon)^2}
\end{equation*}
gives us a second upper bound on the collision probability,
\begin{equation*}
    \mathbb{P}\left[H(f) = H(g)\right] \le P + \frac{\varepsilon r\|f_p\|_{L^{\infty}}}{2(c - \varepsilon)^2}
\end{equation*}
Applying the same trick with the integral
\begin{equation*}
\int_{r/(c+\varepsilon)}^{r/c} f_p(s)\left(1 - \frac{cs}{r}\right) \hspace{0.15cm} ds
\end{equation*}
leads to a second lower bound on the hash collision probability:
\begin{equation*}
    \mathbb{P}\left[H(f) = H(g)\right] \ge P - \frac{\varepsilon r\|f_p\|_{L^{\infty}}}{2(c + \varepsilon)^2}
\end{equation*}

By combining both of the upper bounds and both of the lower bounds, we get the bounds shown in Theorem \ref{thm:approxfun-lphash-prob-bounds}. $\qed$

Note that these are fairly generous bounds -- for instance, $\frac{\varepsilon}{r}\int_0^{r/(c-\varepsilon)} sf_p(s) \hspace{0.15cm} ds$ is generally much less than $\frac{\varepsilon}{c - \varepsilon}$. Nonetheless, they demonstrate that $\Pr[H(f) = H(g)]$ approaches $P$ at a rate of {\it at least} $O(\varepsilon/c)$ or $O(\varepsilon r/c^2)$ as $\varepsilon \to 0$.

\paragraph{Note on choosing $N_f$ and computing $T(f)$} There are two unaddressed issues in our previous discussion: (i) it is unclear how we choose $N_f$ for a function $f(x)$, and (ii) the inner products $\innerprod{e_i}{f}_{L^2_{\mu}}$ may be expensive to calculate, especially if we have to perform some kind of quadrature.

\begin{itemize}
\item[(i)] {\it Choosing $N_f$}: in practice we will combine various heuristics to select a good $N_f$ for which we believe $\hat{f}$ is a good approximation to $f$. For instance, in Section \ref{sec:numerical-experiments} we use Chebyshev polynomials to perform function approximation. Although we choose $N_f = 64$ fixed for demonstration purposes, \citet{Approximation-Theory-Approximation-Practice} and \citet{Chebfun} both describe inequalities and heuristics that can be used to choose a good degree of Chebyshev polynomial (i.e. a good choice of $N_f$) to approximate a function. These bounds are often in terms of approximation in the uniform norm, thus for a bounded domain $\Omega$ give a bound in the $L^2_\mu(\Omega)$ norm.
In the case when $\|f\|_{L^2_{\mu}}$ is known or can be estimated, then $\|\varepsilon_f\|^2_{L^2_{\mu}} = 
\|f\|^2_{L^2_{\mu}} - \|\hat{f}\|^2_{L^2_{\mu}}$ can be explicitly computed (since $\|\hat{f}\|^2_{L^2_{\mu}}=\|T_N(f)\|^2_{\ell^2}$ is computable).
\item[(ii)] {\it Computing $T(f)$}: we will generally not compute $\innerprod{e_i}{f}_{L^2_{\mu}}$ exactly, but rather sample the function at $N$ points and compute some fast unitary transform on those samples to interpolate them by the basis $e_i$. For instance, as part of computing the Chebyshev polynomial coefficients used in Section \ref{sec:numerical-experiments}, we perform a discrete cosine transform on $f$ sampled at certain nodes on the real line. With this approach we don't perfectly extract the coefficients $\innerprod{e_i}{f}_{L^2_{\mu}}$, but we get good approximations to them that improves as $N\to\infty$. For the case of Chebyshev polynomials and smooth functions $f$, this error often reaches very high precision with even moderate $N$ (e.g. $N\approx 100$).
\end{itemize}

\subsection{Monte Carlo methods for function LSH} \label{subsec:monte-carlo-approach}
Our second method for hashing functions generalizes to arbitrary $L_{\mu}^p(\Omega)$ function spaces of finite volume. It comes from the observation that by the theory of Monte Carlo integration,

\begin{align}
    \nonumber \|f - g\|_{L_{\mu}^p} & = \left(\int_{\Omega} \abs{f(x) - g(x)}^p \hspace{0.15cm} d\mu(x)\right)^{1/p} \\
    \nonumber & \approx \left(\frac{V}{N}\sum_{i=1}^N \abs{f(x_i) - g(x_i)}^p\right)^{1/p} \\
    & = \left\|(V/N)^{1/p}\hat{f} - (V/N)^{1/p}\hat{g}\right\|_{\ell^p}
\end{align}

In this expression, $\hat{f} = (f(x_1),\ldots,f(x_n))$ and $\hat{g} = (g(x_1),\ldots,g(x_n))$, and $V= \int_{\Omega} d\mu(x)$ is the volume of $\Omega$. 
The $\{x_i\}_{i=1}^N$ are sampled at random from $\Omega$ under the probability measure $\frac{1}{V}\mu$. It can be shown similarly that $\innerprod{f}{g}_{L^2_{\mu}} \approx \innerprod{(V/N)^{1/2}\hat{f}}{(V/N)^{1/2}\hat{g}}_{\ell^2}$. We can thus view the transform $T(f) = (V/N)^{1/p}\hat{f}$ as an approximate embedding of $L_{\mu}^p(\Omega)$ in $\ell_N^p$.

Naturally, we can extend this idea to the more general class of quasi-Monte Carlo methods to develop other schemes for constructing $\hat{f}$ and $\hat{g}$. For instance, instead of sampling the points $\{x_i\}_{i=1}^N$ i.i.d. under the probability measure $\frac{1}{V}\mu$, we could sample them as a low-discrepancy sequence, e.g. as a Sobol sequence.

\paragraph{Using Monte Carlo to compute hashes} Since the transform $f \mapsto (V/N)^{1/p}\hat{f}$ is an approximate isomorphism between function space and $\ell^p_N$ when $N$ is sufficiently large, we can use many common hash functions for $\ell^p_N$ in $L_{\mu}^p(\Omega)$ by applying them to $(V/N)^{1/p}\hat{f}$. We can summarize the hashing process in three steps:

\begin{enumerate}
    \item Sample $N$ points $\{x_i\}_{i=1}^N$ at random from $\Omega$ (with distribution dependent on the type of Monte Carlo method you wish to apply).
    \item For a similarity of interest on $\ell^2_N$, sample a new hash function $h:\ell^2_N \to \mathbb{Z}$ from relevant LSH family.
    \item When given a new function $f$, sample it at $x_1$ through $x_N$ to generate the vector $\hat{f} = \left(f(x_1),\ldots,f(x_n)\right)$. Apply $h(\cdot)$ to $(V/N)^{1/p}\hat{f}$.
\end{enumerate}

\begin{algorithm}[tb]
   \caption{Function LSH for $L^2$ distance based on \cite{LpHash}, using Monte Carlo methods}
   \label{alg:L2Hash-MonteCarlo}
\begin{algorithmic}
    \STATE {\bfseries Input:} function $f$, sample points $\{x_i\}_{i=1}^N\subseteq\Omega$, coefficients $\{\alpha_i\}_{i=1}^N$, parameters $b$ and $r$
    \STATE {\bfseries Output:} a signed-integer hash $h$
    \STATE
    \STATE $y \gets \left(f(x_1), \ldots, f(x_N)\right)$
    \STATE {\bfseries return} $\left\lfloor \alpha^{\top}y/r + b\right\rfloor$
\end{algorithmic}
\end{algorithm}

\paragraph{Error analysis} Suppose that the points $\{x_i\}_{i=1}^N$ are sampled with distribution $\frac{1}{V}\mu$ from $\Omega$. For sufficiently large $N$, $\frac{V}{N}\|\hat{f} - \hat{g}\|^p_{\ell^p} = \frac{V}{N}\sum_{i=1}^N \abs{f(x_i) - g(x_i)}^p$ is roughly normally distributed via the Law of Large Numbers. This normal distribution has mean $\|f - g\|^p_{L_{\mu}^p}$ and variance
\begin{equation*}
    \text{Var}\bigg(\|T(f) - T(g)\|_{\ell^p_N}^p\bigg) = \frac{V^2}{N}\text{Var}\bigg(\abs{f(x) - g(x)}^p\bigg).
\end{equation*}

Meanwhile, for large $N$ the scaled inner product $\frac{V}{N}\innerprod{\hat{f}}{\hat{g}}_{\ell^2}$ is also approximately normally distributed with mean $\innerprod{f}{g}_{L^2_{\mu}}$ and variance

\begin{equation*}
    \text{Var}\bigg(\innerprod{T(f)}{T(g)}_{\ell^2_N}\bigg)  = \frac{V^2}{N}\text{Var}\bigg(f(x)g(x)\bigg).
\end{equation*}

These equations suggest that our error will be of order $O(N^{-1/2})$. Using quasi-Monte Carlo methods (i.e., by changing our sampling scheme so that we sample from a low-discrepancy sequence), we can achieve an error of $O((\log{N})^dN^{-1})$ \cite{MC-QuasiMC-book} (where $d$ is the dimension of $\Omega$), which may be significantly better than plain Monte Carlo in lower dimensions.

\section{Numerical experiments} \label{sec:numerical-experiments}
To validate the methods described in Section \ref{sec:numerical-experiments}, we ran the following numerical experiments:
\begin{itemize}
    \item measuring hash collision rates for function LSH over cosine similarity;
    \item measuring hash collision rates for function LSH over $L^2$ distance; and
    \item observing the effectiveness of using function LSH for $2$-Wasserstein distance, using the $L^p$ distance formulation of 1D Wasserstein distance in equation \eqref{eq:wasserstein-distance-1d}.
\end{itemize}

We find that in all three experiments, the observed collision rates track closely with the theoretical collision probabilities for the hash function that we are extending from $\ell^2_N$ to $L^2$.

\paragraph{Methodology} For the function approximation method, we used the Chebyshev polynomial basis (which, with a change of variables, can be made a basis for $L^2([a,b])$ with Lebesgue measure). For both methods, we generated 1,024 hash functions in order to measure the average collision probability between a given pair of inputs. We converted each function to a vector in $\mathbb{R}^{64}$ using the two methods described in Section \ref{sec:methods} before hashing them in order to make it easier to compare the effectiveness of both methods. For both methods, this essentially amounts to sampling each function in 64 different locations.

For all experiments, we take $\Omega = [0,1]$. In the second and third experiments, which use the $L^2$-distance hash, we choose the hyperparameter $r$ (from Equation \eqref{eq:lphash-definition}) to be $1$ for demonstration purposes.

\paragraph{LSH over cosine similarity} For our first experiment, we used both of our function hashing methods on pairs of randomly generated sine functions $f(x) = \sin{(2\pi x + \delta_1)}$ and $g(x) = \sin{(2\pi x + \delta_2)}$, where $\delta_1,\delta_2\in \text{Uniform}([0,2\pi])$, since in this parametric form, the true value of $\text{cossim}(f,g)$ can be computed via a closed-form integral.
After converting $f$ and $g$ into vectors in $\mathbb{R}^{64}$ using the two methods described in Section \ref{sec:methods}, we hash them using SimHash \cite{SimHash}, whose collision probability is

\begin{equation}
    \Pr[h(\vec{x}) = h(\vec{y})] = 1 - \frac{1}{\pi}\cos^{-1}\bigg(\text{cossim}(\vec{x},\vec{y})\bigg).
\end{equation}

\begin{figure}
    \centering
    \includegraphics[width=0.5\textwidth]{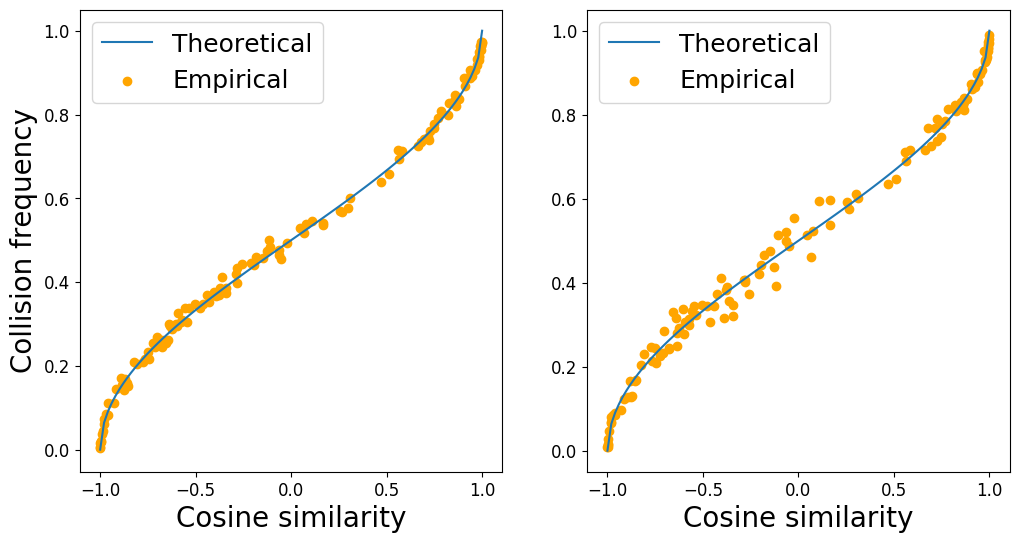}
    \caption{{\it Left:} SimHash observed vs theoretical collision rates using the function approximation method. {\it Right:} observed vs theoretical collision rates using the Monte Carlo approach.}
    \label{fig:simhash-collision-rates}
\end{figure}

This theoretical probability is plotted against the observed collision frequencies in Figure \ref{fig:simhash-collision-rates}.

\begin{figure}
    \centering
    \includegraphics[width=0.5\textwidth]{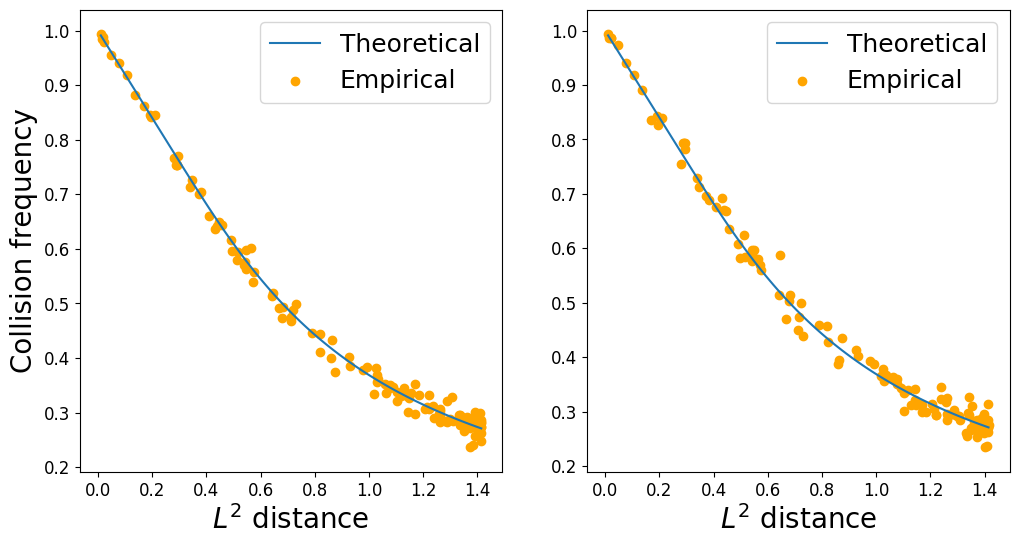}
    \caption{{\it Left}: $L^2$-distance hash observed vs theoretical collision rates using the function approximation method. {\it Right}: observed vs theoretical collision rates using the Monte Carlo approach.}
    \label{fig:L2hash-collision-rates}
\end{figure}

\paragraph{LSH over $L^2$ distance} For our second experiment, we again sample pairs of random sine waves and used the function approximation- and Monte Carlo-based methods to convert the functions to vectors in $\mathbb{R}^{64}$. The collision probability for the $L^2$-distance hash of \citet{LpHash} is

\begin{equation} \label{eq:l2hash-collision-prob}
    \Pr[h(\vec{x}) = h(\vec{y})] = \int_0^r \frac{2}{c\sqrt{2\pi}} e^{-\frac{t^2}{2c^2}} \left(1 - \frac{t}{r}\right) \hspace{0.15cm} dt
\end{equation}

where $c = \|\vec{x} - \vec{y}\|_{\ell^2}$ and $r > 0$ is a user-selected parameter. It follows that when we apply this hash to our vectors in $\mathbb{R}^{64}$, we expect their collision probability to follow the same distribution (except with $\|\vec{x} - \vec{y}\|_{\ell^2}$ replaced by $\|f - g\|_{L^2}$). This is borne out by the observed collision rates shown in Figure \ref{fig:L2hash-collision-rates}.

\begin{figure}
    \centering
    \includegraphics[width=0.5\textwidth]{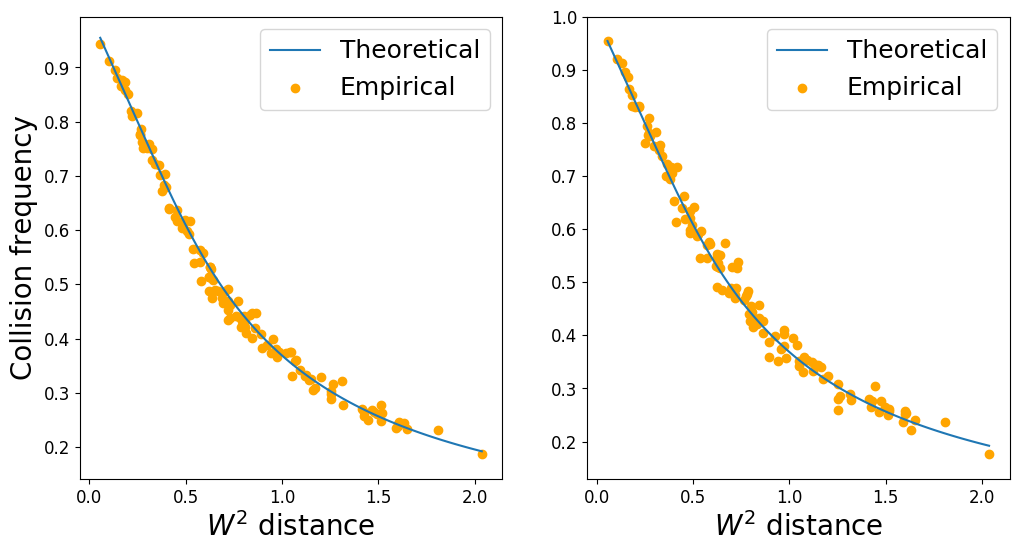}
    \caption{{\it Left:} observed vs theoretical collision rates for the $W^2$-distance hash using the function approximation method. {\it Right}: observed vs theoretical collision rates using the Monte Carlo approach.}
    \label{fig:my_label}
\end{figure}

\paragraph{$2$-Wasserstein distance} For our third experiment, we compare pairs of one-dimensional normal distributions on their second-order Wasserstein distance. We choose to measure the distance between normal distributions because every pair of Gaussians $m_1 = \mathcal{N}(\mu_1,C_1)$ and $m_2 = \mathcal{N}(\mu_2,C_2)$ (with means $\mu_1$ and $\mu_2$ and covariance matrices $C_1$ and $C_2$) has the following convenient closed-form expression for $W^2$ \cite{W2-normal-distribution-simplified}:
\begin{multline*}
    (W^2(m_1,  m_2))^2  = \\
     \|\mu_1 - \mu_2\|_{\ell^2}^2 + \text{tr}\left(C_1 + C_2 - 2(C_2^{1/2}C_1C_2^{1/2})^{1/2}\right).
\end{multline*}
For a pair of 1D Gaussians $m_1 = \mathcal{N}(\mu_1, \sigma_1^2)$ and $m_2 = \mathcal{N}(\mu_2, \sigma_2^2)$, this reduces to
\begin{equation*}
    W^2(m_1, m_2) = \sqrt{(\mu_1 - \mu_2)^2 + (\sigma_1 - \sigma_2)^2}.
\end{equation*}
For our experiment, we repeatedly generated pairs of Gaussians, each with means randomly sampled from $\text{Uniform}([-1,1])$ and variances sampled from $\text{Uniform}([0,1])$. To hash the distributions, we used the expression in Equation \eqref{eq:wasserstein-distance-1d} by hashing\footnote{The inverse c.d.fs are $-\infty$ at $0$ and $+\infty$ at $1$, so we experience some numerical difficulties trying to approximate them by Chebyshev polynomials. To avoid this issue, we only hashed the portion of the inverse c.d.f. living on the interval $[10^{-3}, 1 - 10^{-3}]$ (instead of $[0,1]$), which empirically still performed well in generating a frequency of hash collisions close to the theoretical probability of collision.} the inverse c.d.fs\footnote{A closed-form expression for these inverse c.d.fs does not exist, but this is not an issue because in our experiments we only need to be able to sample these c.d.fs at 64 points in order to hash them.} of the Gaussians on their $L^2$ distance. We then plotted the theoretical probability of collision (Equation \eqref{eq:l2hash-collision-prob}) against the observed frequencies of collisions for both of our methods. These plots are shown in Figure~\ref{fig:my_label}, and show good agreement between predicted collision frequency and empirical collision frequency.

\section{Conclusion}
Similarity search over spaces of functions is a very computationally intensive task. Our study has extended multiple locality-sensitive hash functions from $\ell^p_N$ to the much more general $L^p_{\mu}(\Omega)$ function spaces. These methods can be made arbitrarily precise (i.e., we can get arbitrarily close to the collision probabilities guaranteed by the LSH families in $\ell^p_N$) in exchange for a little more computational effort. From this, the function hashing techniques described in this paper have made the problem of similarity search over function spaces significantly more tractable.

Although we have primarily discussed the cosine similarity hash of \citet{SimHash} and the $L^p$ distance hash of \citet{LpHash}, the methods presented in this paper can in theory be used to extend any hash function for a similarity over $\ell^p_N$ that has an analogous definition in $L^p_{\mu}(\Omega)$. Of particular interest are the hash functions for maximum inner product search of \citet{MIPSHash} and \citet{SignALSH}. Such a hash function could be used as a primitive in defining hash functions for other similarities. For instance, similarity search based on KL divergence can be re-expressed as a maximum inner product search problem, based on the fact that
\begin{align*}
    D_{KL}(p\text{ }\|\text{ }q) & = \int_{\Omega} p(x)\log{\left(\frac{p(x)}{q(x)}\right)} \hspace{0.15cm} dx \\
    & \propto 1 - \frac{1}{\innerprod{p(x)}{\log{p(x)}}_{L^2}}\innerprod{p(x)}{\log{q(x)}}_{L^2}
\end{align*}

where the proportionality coefficient is constant for fixed $p(x)$.

The techniques described in this paper can also be applied in broader input spaces than $L^p_{\mu}(\Omega)$. The function approximation-based approach of Section \ref{subsec:approxfun-approach} can be used to hash any separable Hilbert space in which we have identified an orthonormal basis $\{e_i\}_{i\in\mathbb{N}}$ (or, at the very least, can implicitly compute the inner products $\innerprod{e_i}{f}$). In addition, the Monte Carlo approach can be used on arbitrary sets of $L^p$ functions defined over any finite-volume measure space (including those for which $\Omega \not\subseteq \mathbb{R}^n$, so long as we have a way of sampling functions in this space).

\bibliographystyle{icml2020}

\end{document}